\documentclass[manuscript,screen]{acmart}
\AtBeginDocument{%
  \providecommand\BibTeX{{%
    \normalfont B\kern-0.5em{\scshape i\kern-0.25em b}\kern-0.8em\TeX}}}

\setcopyright{acmcopyright}
\copyrightyear{2022}
\acmYear{2022}
\acmDOI{}

\acmConference[]{}{}{}
\acmBooktitle{}
\acmPrice{}
\acmISBN{}

\usepackage{verbatim} 
\usepackage{tabulary} 
\usepackage[inline]{enumitem} 
\usepackage{xhfill}
\usepackage{makecell} 
\usepackage{todonotes}
\usepackage{lscape}
\usepackage{tabularx}
\usepackage{array}
\usepackage{tablefootnote}
\usepackage{todonotes}
\usepackage{threeparttable}
\usepackage{nomencl}
\makenomenclature

\newcommand{\PreserveBackslash}[1]{\let\temp=\\#1\let\\=\temp}
\newcolumntype{C}[1]{>{\PreserveBackslash\centering}p{#1}}
\newcolumntype{R}[1]{>{\PreserveBackslash\raggedleft}p{#1}}
\newcolumntype{L}[1]{>{\PreserveBackslash\raggedright}p{#1}}

\newcommand{\set}[1]{\mathcal{#1}}
\newcommand{\DSepSet}{\set{W}}
\newcommand{\contexts}{\set{C}}
\newcommand{\context}{C}
\newcommand{\actions}{\set{A}}
\newcommand{\states}{\set{S}}
\newcommand{\dimension}{d}
\newcommand{\discountFactor}{\gamma}

\newcommand{\radius}{rad}
\newcommand{\subgaussian}{\Psi}
\newcommand{\numEp}{N}
\newcommand{\maxarms}{m}

\newcommand{\Alg}[1]{\textsc{#1}}
\makeatletter
\newcommand*{\rom}[1]{\expandafter\@slowromancap\romannumeral #1@}
\makeatother

\newcommand{\Diameter}{D}
\newcommand{\NAgents}{K}
\begin{document}

\title{Survey on Fair Reinforcement Learning: Theory and Practice}

\author{Pratik Gajane}
\email{p.gajane@tue.nl}
\affiliation{%
  \institution{Eindhoven University of Technology}
  \country{The Netherlands}
}

\author{Akrati Saxena}
\email{a.saxena@tue.nl}
\affiliation{%
  \institution{Eindhoven University of Technology}
  \country{The Netherlands}
}

\author{Maryam Tavakol}
\email{m.tavakol@tue.nl}
\affiliation{%
 \institution{Eindhoven University of Technology}
  \country{The Netherlands}
}

\author{George Fletcher}
\email{g.h.l.fletcher@tue.nl}
\affiliation{%
 \institution{Eindhoven University of Technology}
  \country{The Netherlands}
}

\author{Mykola Pechenizkiy}
\email{m.pechenizkiy@tue.nl}
\affiliation{%
 \institution{Eindhoven University of Technology}
  \country{The Netherlands}
}

\renewcommand{\shortauthors}{Gajane et al.}

\begin{abstract}
Fairness-aware learning aims at satisfying various fairness constraints in addition to the usual performance criteria via data-driven machine learning techniques. 
Most of the research in fairness-aware learning employs the setting of fair-supervised learning. However, many dynamic real-world applications can be better modeled using sequential decision-making problems and fair reinforcement learning provides a more suitable alternative for 
addressing these problems. In this article, we provide an extensive overview of fairness approaches that have been implemented via a reinforcement learning (RL) framework. 
We discuss various practical applications in which RL methods have been applied to achieve a fair solution with high accuracy. 
We further include various facets of the theory of fair reinforcement learning, organizing them into single-agent RL, multi-agent RL, long-term fairness via RL, and offline learning. 
Moreover, we highlight a few major issues to explore in order to advance the field of fair-RL, namely -- \         \begin{enumerate*}[label=\roman*)]
    \item correcting societal biases,
    \item feasibility of group fairness or individual fairness, and
    \item explainability in RL.
\end{enumerate*}
Our work is beneficial for both researchers and practitioners as we discuss articles providing mathematical guarantees as well as articles with empirical studies on real-world problems.

\end{abstract}

\ccsdesc[500]{Computing methodologies~Machine learning approaches}
\keywords{Fairness-aware learning, reinforcement learning}

\maketitle

\section{Introduction}




Recently, the machine learning (ML) and artificial intelligence (AI) community have taken major steps in advancing the research towards fairness-aware learning. We point the readers to the following non-exhaustive list of references: \citet{FairnessTextbarocas-hardt-narayanan, FairMLsurveyMehrabi, GajPechFAT2018}. Most of the research has been focused on addressing fairness concerns in supervised learning. 
Let us consider a real-world application of hiring, which we shall use as a running example throughout the article. Consider a firm aiming to hire employees for a number of positions. Common applications of ML in hiring include identifying the candidates, processing incoming applications, and selection (see \citet[Table 4.1]{Krishnakumar2019} and references therein). Typically, ML tools for such scenarios use supervised learning, i.e., they model the relationship between applicants and outcomes from a training dataset, and then apply their model to predict outcomes for subsequent applicants \citep{NBERw27736}. It has been shown that supervised learning techniques can effectively identify high-quality candidates as well as reduce the duration and cost of hiring \citep{NBERw21709}. Nevertheless, supervised learning methods are prone to discrimination or biased outcomes if they are trained on inaccurate \citep{Kim2017}, biased \citep{10.2307/24758720}, or unrepresentative input data \citep{DBLP:journals/corr/abs-1901-10002}. A number of pre-processing and post-processing methods have been proposed to mitigate these issues (e.g., \citep{9517766},\citep{8437807},\citep{NIPS2016_6374}). However, these methods can be expensive, cumbersome, and may also reduce the accuracy of the assessment \cite{RaghavanBrookings2019}. Thus, it is reasonable to look for valid alternatives for fairness-aware supervised learning techniques.  

In addition, it has been shown that imposing fairness constraints as a static singular decision (as standard supervised learning methods do) while ignoring subsequent dynamic feedback can harm sub-populations \citep{pmlr-v80-liu18c, pmlr-v119-creager20a, 10.1145/3351095.3372878}. Hence, it is imperative for researchers and practitioners to account for the indirect and delayed effects of the decisions being made by the ML system. It is customary to model such scenarios as sequential decision-making problems as they allow the consequences of a decision to last for an arbitrarily long time \citep{10.5555/528623}. Indeed, regarding our running example of hiring, \citet{10.5555/3454287.3455691} note: ``Hiring is rarely a single decision point, but rather a cumulative series of small decisions.”

A number of classical optimization methods can be used for sequential decision-making problems; however, they require as input a complete characterization of the learning environment \citep{RLTextSuttonBarto}. In the case of hiring, this translates to a comprehensive knowledge of how hiring decisions affect the goal of a fair system (i.e., identify high-quality candidates while complying with fairness constraints) which, in practice, is not always available. Reinforcement learning (RL) methods provide a suitable solution in such scenarios as they can learn from interacting with an unknown environment. Furthermore, as explained in \citet{RLTextSuttonBarto}, the ability to take into account indirect and delayed consequences of actions is a key feature of reinforcement learning methods. This is because an RL agent's actions are permitted to affect the future state of the environment, thereby affecting the options available to the agent at later times.  

For the task of hiring, supervised learning may also tend to select from groups with proven track records (i.e., \textit{exploitation}), rather than selecting from non-traditional applicants (i.e., \textit{exploration}), raising valid concerns about access to opportunity \citep{NBERw27736}. Even the goal of selecting high-quality candidates requires some exploration in order to possibly discover a better pool of applicants and consequently make better hiring decisions. However, the hiring firm must also exploit its knowledge about the well-understood parts of applicants' pool to make rewarding decisions. Therefore, neither exploration nor exploitation can be pursued exclusively without failing at the task. RL methods are designed to provide a meticulous and goal-directed trade-off between exploration and exploitation, unlike supervised learning methods. Hence, reinforcement learning can be effectively used for real-world applications  (such as hiring, recommendation systems, resource allocation, etc.), posed as a sequential decision-making problem. 

\citet{NBERw27736} provide comparative results of using reinforcement learning and supervised learning in hiring. They focus on the decision to grant first-round interviews for high-skilled positions in consulting, financial analysis, and data science. Using data from professional services recruiting within a Fortune 500 firm, they demonstrate that the approach based on RL improves the quality (as measured by
eventual hiring rates) of candidates selected for an interview while also increasing demographic diversity. On the other hand, supervised learning-based algorithms improve hiring rates but select far fewer Black and Hispanic applicants. Thus, fair reinforcement learning (fair-RL) methods provide a viable solution for fairness-aware decision-making problems which might not be optimally solvable by supervised learning.   

Fair-RL poses additional challenges that traditional fairness-oblivious RL methods are unable to solve. A frequently used performance measure in RL algorithms is the \textit{regret}. Regret of an algorithm measures the difference between the best possible performance and the performance of the said algorithm. Therefore, optimizing the performance is equivalent to minimizing the regret, which is a commonly used goal in fairness-oblivious RL. However, minimizing the regret might not be sufficient for satisfying fairness constraints. It is easy to construct RL solutions that achieve optimum regret while performing unacceptably on fairness requirements. At the same time, dismal regret cannot be tolerated for the sake of a higher level of compliance with the fairness constraints. Hence, fair-RL presents a dual goal of optimizing the usual performance criteria such as regret and adhering to fairness considerations. 

The remainder of this article is structured as follows. In Section \ref{sec:background}, we begin by providing the preliminaries of fair-RL. Then, in Section \ref{sec:AppFairRL}, we furnish the real-world usage of fair-RL methods, followed by an extensive overview of the theory of fair-RL methods in Section \ref{sec:TheoryFairRL}. The kind of problems demonstrated by our running example of hiring mainly fall under single-agent RL, and these methods are enlisted in Section \ref{sec:SingleAgentFairRL}. 
Fair-RL approaches to handle the extensions of our running example to multiple agents, long-term fairness, and offline learning are described in  Section \ref{sec:MultiAgentFairRL}, \ref{sec:LongTermtFairRL}, and \ref{sec:OfflineFairRL} respectively. 
In Section \ref{sec:Challenges}, we discuss a few of the major challenges from fair-RL research, and in the end, we provide brief concluding remarks in Section \ref{sec:conclusion}. 
To the best of our knowledge, the closest work to ours is from \citet{DBLP:journals/corr/abs-2001-04861} that reviews fairness approaches in sequential decision-making problems. While there is a small overlap limited to \citet[Section 3.1 and 4.2.5]{DBLP:journals/corr/abs-2001-04861}, following are the salient additions in our work --
\begin{enumerate*}[label=\roman*)]
\item we include long-term fairness as well as multi-agent RL problems, 
\item we include application of fair-RL approaches to practical problems, and
\item we discuss challenges in fair-RL research at length. 
\end{enumerate*}


\section{Fundamentals of (Fair-)Reinforcement Learning}
\label{sec:background}
In this section, we briefly introduce the fundamental concepts of reinforcement learning. However, providing a thorough review of RL is beyond the scope of this article. We point the readers to \citet{RLTextSuttonBarto, RLTextCsaba} for a thorough understanding of the subject matter. 
Reinforcement learning problems are usually modeled as either multi-armed bandits (MAB), Markov decision processes (MDP), or their variants. Below we briefly introduce some of these formulations. 

\subsection{Multi-Armed Bandit (MAB)} 
\label{sec:IntBandits}

A multi-armed bandit problem can be symbolized as a game from time period $t=1, \dots, T$ between a learner and its environment where $T$ is called the \textit{horizon}. 
\nomenclature{$t$}{Index of a time step}%
\nomenclature{$T$}{Time horizon}%
The learner has a set of arms (or actions, used interchangeably in the rest of the article) $\actions$ 
\nomenclature{$\actions$}{Set of arms/actions}%
\nomenclature{$|\cdot|$}{Cardinality of set $\cdot$}
available to it. At every time period $t = 1, \dots, T$, each arm $a$ is associated with a numerical value i.e., a reward
\footnote{In some scenarios, considering a loss instead of a reward might be more suitable, however, mathematically, the problem settings as well as the general solution strategies are equivalent. In this article, for the sake of uniformity we consider rewards by default and not losses.}.
The learner's task is to pull an arm from $\actions$, whereupon it receives the reward associated with that arm. Such a feedback is called \textit{bandit feedback}\index{bandit feedback}.
The learner's goal is to optimize the value of the arms it chooses, and hence its task is to find a policy that selects the best arm for a given time period. 

In the stochastic version of this problem, a stationary reward distribution is associated with each arm, and the rewards are drawn from the respective distributions. In the adversarial setting, the rewards are generated by an adversary, and the reward probabilities may not be stationary. Generally, in many real-life applications, some information or \textit{context} is available that can be used to make a better decision when choosing amongst all actions. These scenarios are modeled by \textit{contextual bandits}, in which at time step $t$, the learner first observes the context $\context_t$ 
\nomenclature{$\context$}{Context}%
before choosing an action. In the problems described so far, the learner receives absolute feedback about its choices; however, relative feedback is naturally suited to many practical applications where people are expected to provide feedback, e.g., user-perceived product preferences or information retrieval systems. These scenarios are modeled by \textit{duelling bandits} in which the learner is to select two actions at every time step. As feedback, the learner receives information about which action gave a better reward, i.e., won the duel. 
Additionally, in order to produce scalable results when the number of actions is large, the setting of \textit{infinite bandits} is used in the literature. Here, the learner has to choose from a convex set of arms contained in a ball of the given radius. Alternatively, \textit{causal bandits}\citep{10.5555/3157096.3157229} are used to study the problem of using causal models to improve the rate at which good interventions\footnote{We borrow the term from \citet{CausalTextPearl}.} can be learned online.
In a causal bandit problem, interventions are treated as arms, but their influence on the reward — along with any other observations — is assumed to conform to a known causal graph.

Algorithms for MAB problems are typically divided into two groups \citep{kaufmann:tel-01413183} -- 
\begin{enumerate*}[label=\roman*)]
    \item upper confidence bound (UCB) algorithms (frequentist algorithms), and
    \item Thompson sampling based algorithms (Bayesian algorithms). 
\end{enumerate*}  

\subsection{Markov Decision Process (MDP)}
A Markov decision process is characterized by its state set $\states$,
\nomenclature{$\states$}{Set of states}%
the action set $\actions$, and one-step dynamics of the environment, which specify, for any state and action, the probability of each possible pair of next state and reward. The learner's task is to choose an action from $\actions$ at each time step. As with MAB problems, the learner's goal is to optimize the received rewards. In some cases, a reward in the future is deemed not worth quite as much as a reward now. This is expressed by using a discount factor $0 < \discountFactor < 1$
\nomenclature{$\discountFactor$}{Discount factor}%
and a reward that occurs $n$ steps in the future is multiplied by $\discountFactor^n$. In episodic RL problems, an agent interacts with the environment in episodes of fixed length. A partially observable MDP (POMDP) is a generalization of an MDP to model planning under uncertainty. In a POMDP, the learner cannot directly observe the system state, and it uses its observations to form a belief about the current state. 
Q-learning is a popular RL algorithm \citep[Chapter 6]{RLTextSuttonBarto} and deep Q-learning is its variant using deep convolutional neural network \citep{mnih-atari-2013}.  

\subsection{Multi-Agent RL (MARL)} Multi-agent reinforcement learning studies how multiple agents can collectively learn in the same environment. To see how the single-agent RL concepts discussed above can be readily extended to MARL, we point the readers to \citet{MARL_survey}. 

\subsection{Fairness in Reinforcement Learning}
In the literature, fairness in reinforcement learning methods is aimed at either of the following:
\begin{itemize}
    \item Alleviate \textit{societal bias} from the decisions made by learning algorithms (societal fairness) -- \ Societal bias or discrimination in this context refers to unfavorable treatment of people due to their membership to certain demographic groups that are distinguished by the \textit{protected} or \textit{sensitive} attributes, such as race, gender, age, etc. Discrimination, based on protected attributes, is prohibited by international legislation \cite{EuAntidisLaw}. 
     RL techniques have been applied to several real-life problems from different areas to achieve fair solutions, such as hiring \citep{NBERw27736}, disease contagion control problem \cite{atwood2019fair}, face recognition \cite{wang2020mitigating}, recommendation systems \cite{singh2018fairness, burke2018balanced, biega2018equity}, and so on.

    \item Adhere to defined fairness constraints in decision-making or resource allocation (non-societal fairness) -- \ On the other hand, some works frame fairness as a way of adhering to defined constraints in allocation tasks or real-time/sequential decision-making problems. This is particularly common in designing fair solutions for resources allocation schemes in computer networks. Other well known applications include HTTP adaptive streaming \cite{akhshabi2012happens}, traffic management in autonomous vehicles \cite{athalyefairness}, etc. 
\end{itemize}

\section{Real-world Implementations of Fair Reinforcement learning}
\label{sec:AppFairRL}

Fair-RL methods have been applied to achieve fair solutions for various real-world problems. In this section, we discuss some of these works categorized based on societal and non-societal fairness.
\subsection{Implementations of Fair-RL for Societal Fairness}
\citet{NBERw27736} model the decision to extend interview opportunities in the recruitment process as contextual bandits and employ \Alg{UCB-GLM} algorithm \citep{10.5555/3305890.3305895} as a solution. They test the proposed solution on the administrative data of 88,666 job applications made to recruitment services within a Fortune 500 firm. Their results show that the used fair-RL method more than doubles the share of Black and Hispanic
representation in selected applicants. 
At the same time, the quality of the selected applicants (as measured by their hiring potential) is also substantially increased.

\citet{atwood2019fair} study the \textit{precision disease contagion control} problem, which aims to provide a fair allocation of vaccines in society. In such problems, uniform allocation might not be a fair solution as some people might be at more risk due to their characteristics, such as age, or due to their position in the network. A fairness-aware vaccine allocation policy aims at finding allocation strategies that equalize outcomes in the population. They train a Deep Q-Network \citep{mnih2015human} using dopamine's experiment runner and assume that the agents providing the vaccines can observe the contact network structure of individuals and their disease state. The reward function at each step is the negative number of newly sick nodes. 
The results show that the proposed solution achieves fairness as compared to baselines. These promising results confirm that RL can be further used to achieve fair solutions for such social problems having complex network structures. 

\citet{wang2020mitigating} propose an RL-based race balance network (RL-RBN) that reduces the skewness of features between different races for mitigating the bias in face recognition. An MDP-based setting is used to find the optimal margins for non-Caucasians, and then a deep Q-learning method is applied to learn policies for an agent to select an appropriate margin.
Fairness is evaluated using standard deviation and skewed error ratio (SER), and the results show the effectiveness of RL-RBN.

\subsection{Implementations of Fair-RL for Non-societal Fairness}
Another well-known application of fairness-aware solutions using RL is \textit{Enhancing transmission control protocol} (TCP) over wireless mesh networks (WMN). 
Generally, TCP is not fair in resource allocation over WMN \citep{franceschinis2005measuring}. \citet{arianpoo2016network} propose a distributed mechanism that observes the unfairness in resource allocation and then tunes the TCP parameters accordingly. In their solution, each TCP source models the state of the multi-hop network as an MDP, and then they use the Q-learning algorithm to monitor and learn the transition matrix of the proposed MDP. The agent learns the network usage behavior of each node using the local fairness index and the aggressive index. The reward function is used for guiding the learning agent to choose the correct actions that will eventually provide a fair solution in a distributed manner. The proposed solution improves the fairness of the flows traversing a larger number of hops with a negligible impact on the smaller size flows by tuning TCP parameters. The overall performance is enhanced by 10-20\% without using feedback messaging and no extra overhead to the system. 
\citet{yamazaki2021fairness} present an RL-based method, called QTCP-AIMD, that uses Q-learning based TCP with Additive Increase Multiplicative Decrease (AIMD) to improve fairness in congestion window control mechanism. The simulation analysis shows that the proposed method improves fairness without degrading both throughput as well as low latency characteristics of QTCP. \citet{tong2021qoe} propose a deep RL-based method for dynamic spectrum allocation in the case of resources shortage. 

\textit{HTTP adaptive streaming} (HAS) is a popular service for adaptive video streaming, in which each video is temporally segmented and stored in varying levels of quality. The quality selection heuristics used by the player dynamically request for the most appropriate quality level based on the network conditions. Most of these heuristic methods are fixed and deterministic, and are not able to provide good performance if the network conditions are highly dynamic. The quality of experience for users might be poor due to freezing or frequent quality switches while playing the videos. In real scenarios, multiple clients share a single medium and request videos from the HAS server. The mutual synchronization among clients gives rise to unfairness among them as one client might have a negative impact on others \citep{akhshabi2012happens}. \citet{petrangeli2014multi} propose a multi-agent Q-learning based HAS client that will learn and dynamically adapt its behavior depending on the network conditions to achieve fairness and provide a high quality of experience which is measured using average mean opinion score \cite{de2013model}. Some other works on fairness-aware resource optimization include \citet{chen2021bringing} and \citet{ valkanis2022efficiency}.

RL-based solution for autonomous vehicles by adding fairness terms in the reward function significantly improves the learned behavior of the agents to avoid bottleneck situation \citep{athalyefairness}. The results show that an RL method provides high outflow (vehicles per hour) and throughput efficiency \cite{vinitsky2018benchmarks}. The authors study a single-agent method, and further investigation in multi-agent setting would be interesting. Applying fairness penalties in multi-agent settings for other non-bottleneck situations, such as changing lanes or speed unnecessarily, are also open research questions.

RL has provided promising results to achieve fairness and quality of service in \textit{radio resource management} by selecting scheduling rules to allocate users' data packets in the frequency domain. In \citet{comsa2012novel}, Q-learning is used to learn different policies for scheduling rules at each transmission time interval. The proposed solution achieves different levels of throughput-fairness trade-off by offering optimal solutions according to the channel quality indicator for different classes of users. 
\citet{comcsa2014scheduling} propose an RL-based technique that considers the traffic load and channel conditions to learn the parameters for a Generalized Proportional Fair scheduling rule that respects the fairness criteria. The authors further propose a more complex RL framework to achieve higher NGMN (Next Generation of Mobile Networks) fairness \citep{comcsa2014adaptive, comsa2019comparison}. 
\citet{yuan2020fairness} model the \textit{resource block group} allocation problem as a stochastic game framework and provide a multi-agent RL-based solution to achieve fairness by optimizing 5-percentile user data rate. 

In the past few years, researchers have focused on designing fairness-aware recommendation systems \citep{singh2018fairness, burke2018balanced, biega2018equity}. Apart from recommendation systems based on static settings, RL-based systems have provided good performance for interactive systems where users' evolving preferences must be taken into account while determining if the system is fair \citep{zhang2021recommendation}. \citet{liu2020balancing} introduce an RL-based framework, called FairRec, that maximizes the cumulative reward function based on both fairness and accuracy, and maintains a dynamic balance between both in the long run for an interactive recommendation. The fairness is evaluated using Unit Fairness Gain, which considers both a higher conversion rate and a higher weighted proportional fairness.

Furthermore, human-in-the-loop IoT systems are popular means to provide a personalized experience. These systems continuously learn human behavior and evolve to adapt to the human and environment state and take actions autonomously or by way of recommendation. A major challenge in designing personalized IoT applications arises due to human variability as different people interact with IoT applications in different ways or change their behavior over time. \citet{elmalaki2021fair} propose Fair-IoT using a reinforcement learning-based framework that continuously monitors the human state and changes in the environment to adapt its behavior accordingly. The proposed adaptive and fairness-aware solution is further used for two human-in-the-loop IoT applications --  
\begin{enumerate*}[label=\roman*)]
\item automotive advanced driver assistance systems, and 
\item smart house.
\end{enumerate*}
 The analysis shows that the proposed solution improves systems' fairness by 1.5 times and enhances the human experience by 40\% to 60\% compared to non-personalized systems.  

RL-based techniques have been employed to find fair solutions in other domains as well, such as reducing the air traffic congestion \citep{de2010fairness}, traffic light control to optimize the waiting time of all the drivers \citep{li2020fairness, raeis2021deep}, resources management \citep{liu2018energy,qi2020energy,arani2021fairness}, electric taxi charging \citep{wang2021data}, resource allocation in robot systems (human-robot systems, multi-robot systems) \citep{zhu2018deep, claure2019reinforcement}, and same-day delivery services \citep{chen2020same}.

\subsection{Unexplored Issues in the Implementations of Fair-RL}
In this section, we summarized the fair-RL methods that have been deployed in the real-world. Following are a few of the major unexplored issues concerning them. 
\begin{enumerate}
    \item Interpretability/Explainability: Many of the above works use deep RL methods and it remains difficult to diagnose what aspects of the input affect the decisions of these methods. Explainability in deep RL methods is still not widely studied \citep{DBLP:journals/corr/abs-2004-14545}. The lack of comprehensible explainability of automated methods is a concern in many real-world domains such as legislation, law enforcement, healthcare, etc. 
    As explained in \citet{corbettdavies2018measure}, an opaque system may engender mistrust from policymakers and stakeholders thus hindering implementation. 
    In particular, the need for explainability for algorithmic hiring is asserted by \citet{10.5555/3398761.3398960}. This point is further discussed in detail in Section \ref{sec:ExpRL}.
    \item Focus on societal-fairness: As we shall see in Section \ref{sec:TheoryFairRL}, most of the research in the theory of fair-RL is motivated by societal fairness. As witnessed in the current section, practical use of fair-RL lags behind the theory in this aspect. In addition to being a legal requirement in many parts of the world, satisfying a valid societal fairness constraint is beneficial from a business perspective too. 
    Studies demonstrate that if a lack of fairness or equality is perceived by people, it might impact their job satisfaction \cite{McFarlin1992}, which results in retaliatory behavior from the affected parties \cite{Skarlicki1997}.
    We have already noted how fair-RL methods fare better than traditionally used fair-supervised learning methods for tasks like hiring \citep{NBERw27736}.  
    We believe business needs and evidence of the viability of fair-RL methods will naturally motivate extensive use of fair-RL techniques for societal fairness in the real world. 
    
    \item Cold start: RL methods sometimes tend to suffer from ``cold start" i.e., poor performance at the beginning. This might hinder implementation of fair-RL methods in critical domains. Some techniques (e.g., \citep{DBLP:journals/corr/abs-1910-09986}) have been suggested to overcome this problem and it needs to be studied if they can be incorporated in fair-RL methods too. 
\end{enumerate}

\section{Theory of Fair Reinforcement Learning}
\label{sec:TheoryFairRL}
 We divide this section into multiple subsections based on the type of the used RL method. We elucidate some articles in detail while only briefly mentioning others and deferring their details to appendix \ref{Appd:TheoryFairRL} due to space constraints.

\subsection{Fairness in Single-Agent RL}
\label{sec:SingleAgentFairRL}
An extensive number of fair-RL approaches have been developed in single-agent settings. We categorize them according to the used fairness notion for ease of exposition. Table \ref{tab:Performance bounds} provides a comparative view of some of the noteworthy theoretical results. Lack of space prevents us from including a definition of each symbol in the table. Please refer to the description in the relevant text or in the list of symbols given in Appendix \ref{Appd:SymbolTable}.

\begin{table*}
  \caption{Performance bounds}
 \begin{tabular}{p{2.1cm}C{2cm}C{2.4cm}C{3.2cm}C{3.4cm}}
    \toprule
	\textbf{Reference} 
	& \textbf{Formulation} 
	& \textbf{Fairness Notion} 
	& \textbf{Performance Measure} 
	& \textbf{Bound} \\ 
    \midrule
    \citet[Theorem 2]{NIPS2016_eb163727} 
    & Stochastic bandits 
    & Individual fairness 
    & Cumulative regret 
    & $\tilde{O}\left(\sqrt{|\set{A}|^3 T}\right)$ \\
    \hline
    \citet[Theorem 1]{DBLP:journals/corr/JosephKMNR16} 
   & Contextual bandits 
   &  Individual fairness 
   & Cumulative regret   
   &  $\tilde{O}(\dimension |\set{A}|^2\sqrt{T})$ \\ 
   \hline
   \citet[Theorem 2]{DBLP:journals/corr/JosephKMNR16} 
   & Infinite bandits 
   &  Individual fairness 
   & Cumulative regret   
   & $\tilde{O}\left(\frac{\radius^6 \subgaussian^2}{\kappa^2\lambda^2 \Delta_{gap}^2} \right)$  \\
   \hline
    \citet[Theorem 6]{pmlr-v70-jabbari17a} 
   & Discounted reward MDP 
   & Individual fairness
   & Sample complexity to achieve (near-)optimality 
   & $\tilde{O}\left( \frac{f |\set{S}|^5 |\set{A}|^{\frac{1}{1-\gamma}+5}}{(1- \gamma)^{12}} \right)$  \\
   \hline
   \citet[Theorem 4.1]{1112306} 
   & Stochastic bandits 
   & Individual fairness 
   & Expected cumulative deviation from the fairness constraint 
   & $\tilde{O}\left((|\set{A}|T)^{2/3}\right)$ \\
   \hline
    \citet[Theorem 5.2]{1112306} 
   & Dueling bandits 
   & Individual fairness 
   & Expected cumulative deviation from the fairness constraint 
   & $\tilde{O}\left(|\set{A}|^{4/3}T^{2/3}\right)$ \\
   \hline
    \citet[Theorem 3]{10.5555/3327144.3327185} 
   & Linear contextual bandits 
   & Individual fairness 
   & Cumulative regret 
   & $\tilde{O}\left( |\set{A}|^2 \dimension^2 \log{(T)} + \dimension \sqrt{T} \right)$ \\
   \hline 
    \citet[Corollary 4.1]{DBLP:journals/corr/abs-2010-12102}
    & Contextual bandits
    & Group fairness
    & Cumulative regret
    & $O\left( \dimension\sqrt{T} \log{(TL)}\right)$ \\
    \hline 
    \citet[Theorem 5.1]{pmlr-v130-wen21a} 
    & Episodic discounted MDP 
    & Group fairness
    & Cumulative regret 
    & $O\left( \left( N^{2/3} + 1/\epsilon^2 \right) \log{(1/\delta)}\right)$  \\
   \hline
    \citet[Theorem 1 \& 2]{10.1145/3319502.3374806} 
    & Stochastic bandits
    & Minimum selection criteria 
    & Cumulative regret
    & $O\left(\sqrt{(|\actions|T \log{T})} + |\actions|\log{T} \right)$ \\
    \hline
    \citet[Theorem 2]{8737461} 
    & Comb. sleeping bandits 
    & Minimum selection criteria
    & Time-average regret 
    & $ O\left( \frac{|\set{A}|}{2\eta} + \frac{\sqrt{\maxarms |\actions|T \log{T}} + |\actions|}{T} \right)$ \\
    \hline
    \citet[Theorem 8]{JMLR:v22:20-704} 
    & Stochastic bandits 
    & Minimum selection criteria
    & Cumulative regret
    & $O(\sqrt{T\log{T}})$  \\
    \hline
    \citet{10.5555/3398761.3398990} 
    & Contextual bandits 
    & Minimum selection criteria
    & Cumulative regret
    & $O\left(\sqrt{T |\contexts| |\set{A}| \log(|\set{A}|)}\right)$ \\ 
   \hline
    \citet[Theorem 1]{10.1145/3287560.3287601}
    & Stochastic Bandits
    & Selection within pre-specified range
    & Cumulative regret
    & $O\left(\frac{|\actions|}{\Delta_{gap}^2}\log(T)\right)$ \\
    \hline
    \citet[Theorem 3.2.2, 3.3.2]{pmlr-v139-wang21b} 
    & Stochastic bandits 
    & Selection proportional to merit 
    & Cumulative regret 
    &  $\tilde{O}\left(\sqrt{|\actions|T}\right)$ \\
   \hline
   \citet[Theorem 4.2.2, 4.3.2]{pmlr-v139-wang21b} 
   & Linear stochastic bandits 
   & Selection proportional to merit  
   & Cumulative regret 
   & $\tilde{O}\left( \dimension \sqrt{T}\right)$ \\
   \hline
   \citet[Theorem 3]{huang2021achieving} 
   & Causal bandits 
   & Counterfactual individual fairness 
   & Cumulative regret 
   & $\tilde{O}\left( \frac{\sqrt{|\DSepSet|T}}{\tau- \Delta} \right)$\\
    \bottomrule
\end{tabular}
\label{tab:Performance bounds}
\end{table*}

\subsubsection{Individual fairness in single-agent RL}
A commonly used fairness notion is individual fairness \cite{Dwork:2012:FTA:2090236.2090255} which stipulates an RL system to make similar decisions for similar individuals. 

\citet{NIPS2016_eb163727} consider a notion of individual fairness called \textit{meritocratic fairness} in stochastic bandits and contextual bandits. This fairness notion stipulates that a fair algorithm should not select one arm over another if the chosen arm has a lower expected reward than the unchosen arm. 
They provide a lower bound, which proves that no algorithm has diminishing regret before $\Omega(|\set{A}|^3)$ time steps. For stochastic bandits, the authors propose 
an algorithm called \Alg{FairBandits} 
and prove a $\tilde{O}\left(\sqrt{|\set{A}|^3 T}\right)$ upper bound on its regret which matches the lower bound when $T \approx |\set{A}|^3$. For any contextual bandit problem, the authors demonstrate that the optimal learning rate of any fair algorithm is determined by the best KWIK (``Knows When It Knows") \cite{Li2011} bound for the problem. 
\citet{NIPS2016_eb163727} also provide a reduction from KWIK learning algorithm to a fair contextual bandit algorithm and vice versa. 

\citet{pmlr-v70-jabbari17a} extend meritocratic fairness to MDPs with discounted rewards. 
Their fairness constraint called, approximate-action fairness,  requires that an algorithm never favors an action of substantially lower quality over a better action. They design an algorithm, called \textit{Fair-E$^3$}, satisfying approximate-action fairness which achieves (near-)optimality after at most $\tilde{O}\left( \tfrac{f |\set{S}|^5 |\set{A}|^{\frac{1}{1-\discountFactor}+5}}{(1- \discountFactor)^{12}} \right)$ steps where $f$ is a problem-dependent value and $\discountFactor$ is the discount factor.
\nomenclature{$f$}{Problem-dependent value from \citet{pmlr-v70-jabbari17a}}%
%

It has been argued in the literature (e.g., \cite{1112306}) that meritocratic fairness 
suffers from the following limitations -- 
\begin{enumerate*}[label=\roman*)]
\item \label{bla} it allows a subset of arms best in expectation by only a small margin to be selected all the time, even if any single sample from the subset may be worse than a single sample from another subset,
\item it does not constrain the learner in case one subset of arms is much better than other subsets,   
\item it does not account for/correct past inequities or inaccurate/biased data, and 
\item it assumes an accurate mapping from features/arms/actions to true quality is available for the task at hand.
\end{enumerate*}

\citet{1112306} impose the following constraints on stochastic bandits and dueling bandits: 
\begin{enumerate*}[label=\roman*)]
    \item \textit{smooth fairness} constraint - two arms with similar distribution should be selected with similar probability, and 
    \item \textit{calibrated fairness} constraint - sample each arm with probability equal to its reward being the greatest.
\end{enumerate*}
They further elucidate how these two constraints address the first two critiques of meritocratic fairness presented earlier. 
\citet{DBLP:journals/corr/JosephKMNR16} extend upon \citet{NIPS2016_eb163727} by addressing a couple of shortcomings of the latter. Firstly, the problem setting by \citet{DBLP:journals/corr/JosephKMNR16} allows for multiple individuals per group per round, and the learner can choose multiple individuals per round, unlike that of \citet{NIPS2016_eb163727}. 
\citet{10.5555/3327144.3327185} contend that similarity-metric based fairness definitions suffer from the problem that it may be actually difficult for anyone to precisely express a quantitative metric over individuals. To circumnavigate this problem, they assume that the algorithm has access to an oracle that can ascertain what it means to be fair, but cannot explicitly enunciate the fairness metric. After the learner makes its decisions, it receives feedback from the oracle about violations of the fairness constraint.  
For more details about \citet{1112306}, \citet{DBLP:journals/corr/JosephKMNR16}, and \citet{10.5555/3327144.3327185}, please refer to Appendix \ref{Appd:TheoryFairRL}. 
\subsubsection{Group fairness in single-agent RL}
Another common notion is group fairness which imposes statistical/demographic parity in the decisions made by an RL system.

\citet{DBLP:journals/corr/abs-2010-12102} consider group fairness in contextual bandits focusing on the practical application of recommendation systems.
Group fairness constraint here necessitates the expected mean reward of the protected group and that of the unprotected group to be equal (barring a small additive tolerance degree). 
The considered performance measure is expected cumulative regret compared to the best algorithm satisfying the fairness constraint. They further propose an algorithm called \Alg{Fair-LinUCB}
, and prove an upper bound of $O\left( \dimension\sqrt{T} \log{(TL)}\right)$, where $\dimension$ is the dimension of the feature vector and $L$ is an upper bound on the $L2$ norm of all the feature vectors. 
\nomenclature{$L$}{Upper bound on the $L2$ norm of all the feature vectors \citep{DBLP:journals/corr/abs-2010-12102}}%
The experimental results on simulated datasets demonstrate that \Alg{Fair-LinUCB} achieves competitive regret while complying with the considered fairness notion. 

\citet{DBLP:journals/corr/abs-1912-03802} study contextual bandits with the following two definitions of group fairness -- 
\begin{enumerate*}[label=\roman*)]
\item demographic parity, wherein the probability of choosing a group is the same across groups, and
\item proportional parity, wherein the probability of choosing an arm from a particular group is proportional to the size of that group.
\end{enumerate*}
\citet{pmlr-v130-wen21a} develop fair decision-making policies in discounted MDPs, and their approach works with demographic parity and equal opportunity \citep{NIPS2016_6374}.   
For more details about \citet{DBLP:journals/corr/abs-1912-03802} and \citet{pmlr-v130-wen21a}, please refer to Appendix \ref{Appd:TheoryFairRL}.

\subsubsection{Minimum selection criteria}
\citet{10.1145/3287560.3287601} study stochastic bandits considering the practical application of preventing polarization in personalized online spaces \citep{10.1145/3178876.3186143,pmlr-v81-speicher18a}. 
In this setting, the fairness constraint demands that the 
probability 
with which an algorithm selects a group of arms stays within a pre-specified interval. 
They propose an algorithm called \Alg{Constrained-$\epsilon$-Greedy} 
and prove an upper bound of $O\left(\frac{|\actions|}{\Delta_{gap}^2}\log(T)\right)$ on the expected cumulative regret where $\Delta_{gap}$ is the difference between the maximum and the second maximum expected rewards.
\nomenclature{$\Delta_{gap}$}{Difference between the maximum and the second maximum expected rewards in \citet{10.1145/3287560.3287601}}%
Additionally, they provide empirical results showing competent performance of their algorithm on a dataset of news articles with the aim of diversifying across topics, and the MovieLens dataset \cite{10.1145/2827872} with the aim of diversifying recommendations. 

\citet{10.1145/3319502.3374806} explore stochastic bandits with a fairness constraint that a minimum pulling rate for each arm is satisfied (in expectation or anytime throughout the time horizon). 
The performance of the proposed algorithms is measured in terms of regret with respect to the optimal benchmark strategies satisfying the fairness constraint. The authors provide UCB-based algorithms to satisfy both the flavors of the above fairness constraint and prove a regret bound of $O\left(\sqrt{(|\actions|T \log{T})} + |\actions|\log{T} \right)$. 

\citet{8737461} also employ minimum selection criteria for each individual arm in a setting called combinatorial sleeping bandits in which multiple arms (up to $\maxarms$)
\nomenclature{$\maxarms$}{Maximum number of arms that can be played simultaneously in \citet{8737461}}%
can be simultaneously played, and an arm could sometimes be ``sleeping" (i.e., unavailable). The authors enlist real-time scheduling in wireless networks, ad placement in online advertising systems, and task assignment in crowd-sourcing platforms as practical applications and present experimental results on simulations showing the competent performance of their approach.
\citet{JMLR:v22:20-704} consider stochastic bandits where the minimum selection fraction of each arm is limited to $\left[0,\frac{1}{|\set{A}| -1}\right)$. 
\citet{10.5555/3398761.3398990} investigate the problem of an AI system assigning tasks or distributing resources to multiple humans. They model this problem using the setup of contextual bandits, and their fairness criterion demands a minimum rate for assigning a task or a resource to an individual, which translates to a minimum probability $\in \left(0, \frac{1}{|\set{A}|}\right)$ of each arm being pulled. 
For more details about \citet{8737461}, \citet{JMLR:v22:20-704}, and \citet{10.5555/3398761.3398990}, please refer to Appendix \ref{Appd:TheoryFairRL}.

\subsubsection{Other notions}
\citet{pmlr-v139-wang21b} aim to achieve merit-based fairness of exposure to the items while optimizing utility to the users in stochastic bandits and stochastic linear bandits. The fairness constraint requires that each arm receives an amount of exposure proportional to its merit, where merit is quantified through an application-dependent merit function.
The experimental results on synthetic data and on real-world data from ``Yahoo! Today" Module \citep{10.1145/1772690.1772758} show that their proposed algorithms provide better fairness in terms of exposure to the items. 
\citet{huang2021achieving} study counterfactual individual fairness in causal bandits. They focus on the practical application of online recommendations and aim to provide user-side fairness for customers.
Experimental results provided on the Email Campaign data \citep{lu20regret} show that their proposed algorithm maintains good performance while satisfying counterfactual individual fairness in each round. 
\citet{DBLP:journals/corr/abs-2111-14387} provide an RL solution for fair and efficient allocation of a divisible resource among a population with distinct preferences. They model this problem using adversarial bandits and propose an algorithm based on \Alg{EXP3} \citep{10.1137/S0097539701398375} which attains sub-linear bounds for both cumulative regret and fairness-regret if the adversary is restrained in a certain way.
For more details about \citet{pmlr-v139-wang21b}, \citet{huang2021achieving}, and \citet{DBLP:journals/corr/abs-2111-14387}, please refer to Appendix \ref{Appd:TheoryFairRL}.

\citet{nabi2019learning} study fairness in RL from the perspective of causal inference and constrained optimization. They present several strategies for learning optimal
policies by modifying some of the existing RL algorithms, such as Q-learning, which also account for some fairness considerations, and further provide a theoretical guarantee that their proposed fair policy satisfies the specified fairness constraints.

\citet{10.1145/3224431} address the problem of providing \textit{proportionally fair} allocations in a sequential resource allocation problem where a set of tasks are to be handled by some servers. In their work, a learner is considered fair if all the tasks are served at similar rates, avoiding starvation. In a proportionally fair allocation, the aim is to maximize the sum of the logarithm of the task expected service rates.



\subsection{Fairness in Multi-Agent RL}
\label{sec:MultiAgentFairRL}
Balancing fairness is essential in many multi-agent systems in which fairness constraints must be satisfied among several users simultaneously. In a modification of our running example, suppose multiple divisions/subsidiaries of the same employer are looking to hire for a number of positions. Here, the hiring process of each subsidiary can be thought of as a single agent. A fair-RL algorithm for this problem aims for a fair solution for both the employer  -- to have a desirable set of employees, and applicants – to receive decisions that are fair (non-discriminatory) with respect to the protected attributes. 

Fairness in multi-agent systems has been studied in reinforcement learning scenarios \citep{peysakhovich2018prosocial}. 
In one of the early attempts, \citet{zhang2014fairness} model fairness in multi-agent sequential decision-making via a simple linear programming approach. The fairness optimization is then formulated in a game-theoretic framework to achieve a Nash equilibrium that contains a regularized max-min fairness policy.  
\citet{zhu2018deep} propose a multi-type resource allocation method for multi-robot systems by leveraging a weighted combination of utility and inequality, which is measured using the Gini-index, as the overall objective function in their approach. Moreover, \citet{agarwal2019reinforcement} explore multi-agent settings with non-Markovian reward functions and develop model-based and model-free algorithms for joint decision making of multiple agents. 
They present a tabular model-based algorithm using Dirichlet sampling to obtain a regret bound of $O\left(\NAgents \Diameter |\states||\actions|\sqrt{\frac{|\actions|}{T}}\right)$, where $\NAgents$ is the number of agents and $\Diameter$ is the diameter of the underlying Markov chain.
\nomenclature{$\NAgents$}{Number of agents in \citet{zhu2018deep}}%
\nomenclature{$\Diameter$}{Diameter of the Markov Chain in \citet{zhu2018deep}}%
In a more decentralized setting, \citet{jiang2019learning} introduce a hierarchical approach for joint policy optimization. In this setting, each agent first learns its own policy to balance fairness versus efficiency, and at a higher level, a controller maximizes the multi-objective reward by switching between the local policies while interacting with the environment. In their approach, the fairness measure is decomposed between the agents such that they only focus on optimizing their corresponding sub-goals. Additionally, \citet{zimmer2021learning} study cooperative multi-agent scenarios and present a fair policy learning to optimize a welfare function that utilizes both fairness efficiency and fairness equity in a decentralized manner. Their framework consists of a self-oriented and a team-oriented network which are together optimized via a policy gradient algorithm with theoretical proof of convergence.  

Furthermore, \citet{wang2019evolving} specifically show that a model which only focuses on maximizing the team reward might lead to a conflict and potential unfair outcomes for individual members. Accordingly, they develop a modular architecture to improve cooperation in social dilemmas. The proposed architecture adapts a fast timescale learning of individual agents combined with a slow evolution mechanism that allows for a natural selection in a population. \citet{grupen2021fairness} define a group-based measure of fairness for fully cooperative multi-agent settings and present an approach in which the agents learn how to coordinate fairly toward their common goal. They benefit from equivariant policy learning to achieve provably fair outcomes for individual members of a team where the trade-off between fairness and utility is obtained dynamically.  
Multi-agent RL techniques are also employed in other applications, such as learning stock trading strategies to keep a balance between revenue and fairness for the involved
clients \citep{bao2019fairness}.

\subsection{Long-term Fairness via RL}
\label{sec:LongTermtFairRL}


Most approaches designed to deal with fairness-aware learning only consider the immediate implication of bias in a static context, whereas in many applications, the decisions made by RL systems have consequences in the long term. Recent work explores the long-term effects of RL and demonstrates that modeling the immediate effect of decisions for single-step prevention of bias does not guarantee fairness in later downstream tasks \citep{kannan2019downstream,milli2019social,liu2018delayed}. For example, \citet{Holzer2007} explore the long-term effects of affirmative action \citep{deshpande2005affirmative} in hiring. They cite several studies to demonstrate that there is little evidence for affirmative action leading to weaker performance. Moreover, it clearly generates positive benefits for the minority and low-income communities, and perhaps for employers as well. They also enlist positive indirect consequences of affirmative action in certain sectors -- e.g., medical care, where minority physicians are more likely to provide medical care to minorities and low-income communities. Fairness methods in RL aiming to deal with such long term effects are presented below. 

 \citet{kannan2019downstream} study two-step decision scenarios and illustrate that a fair decision of a classifier in the first step has an indirect unfair impact on the next Bayesian decision-making task. \citet{milli2019social} introduce robust strategies to address a similar problem in the context of strategic classification, which aims at lowering the social burden and disadvantage in certain groups in the long-term. \citet{liu2018delayed} propose a one-step feedback model that reveals how decisions change the underlying population over time, confirming that common fairness criteria, in general, do not lead to reshaping the population and promoting long-term improvements. 

The consequences of bias can go beyond the immediate next step and impact the stakeholders further steps away. In such scenarios, reinforcement learning is a remedy to model the long-term implications of bias in the form of Markov decision processes. \citet{d2020fairness} discuss the growing need to understand the long-term behaviors of deployed ML-based decision systems and their potential consequences, and accordingly, propose a framework for systematically exploring these long-term effects. \citet{tu2020fair} study the dynamics of population qualification and algorithmic decisions under a partially observed MDP and show that the long-term effects are heavily shaped by the interplay between algorithmic decisions and individuals' reactions. They further illustrate that the same fairness constraint can have an opposite impact depending on the underlying problem scenarios, which highlights the importance of understanding real-world dynamics in decision-making systems. These approaches are solely based on simulations by reproducing the dynamics of certain applications with known parameters to highlight the need for modeling the long-term implications of bias. Subsequently, they have been employed in some applications such as recommendation systems to retain fairness for a longer period of time \citep{singh2020building,zhang2021recommendation}. 

Consequently, we require methods that can learn the dynamics of a population and the optimal policies which are unbiased in the long-term. \citet{wen2021algorithms} explore the temporal effects of fair/unfair decisions for every individual in the population. They first formulate fairness definitions in an MDP setting and then propose 
an algorithm to learn an optimal policy that satisfies the fairness constraints. The presented experimental results show that accounting for the dynamic effects of decisions improves the results of supervised learning under optimistic assumptions. \citet{siddique2020learning} formulate a fair optimization problem in a multi-objective sequential decision-making setting. They propose to adjust the standard reward function by applying a social welfare function on the reward distribution, leading to a multi-objective reinforcement learning problem based on welfare optimization. All these works underline the importance of modeling long-term fairness, which is still in the beginning phase, and more research needs to be done in the domain of fairness in reinforcement learning to account for the dynamic effects of decisions in our society.%

\subsection{Offline Fair-RL}
 \label{sec:OfflineFairRL}

Consider another variant of our running example: hiring via predictive analysis of offline data using learning models \citep{PESSACH2020113290}. In such scenarios, historical recruitment data is used to make predictions about future hiring activities and candidates. Such problems are solved using offline fair-RL methods.

\citet{metevier2019offline} present a theoretically grounded contextual bandit algorithm for offline data.
Their approach can adapt to various definitions of fairness and returns ``no solution found" if data is insufficient or there might be conflict. 
In addition to RL and standard bandit settings, fairness-aware learning has been employed in counterfactual learning frameworks for scenarios that can only be evaluated on the historical data, and where online learning/interaction is infeasible, costly, or dangerous, e.g., in healthcare applications. \citet{tavakol2020fair} propose to model the task of fair classification as a bandit problem in which group-based fairness constraints can be satisfied via off-policy learning in a counterfactual bandit setting. Consequently, counterfactual risk minimization techniques are leveraged to learn fair policies from biased offline data. In another work, \citet{coston2020counterfactual} present a counterfactual risk assessment method using doubly robust estimation. This method reduces the likelihood of an adverse event by identifying risky cases where the decision-making needs human intervention. They further demonstrate that modeling the counterfactual outcomes is necessary in order to effectively conduct fairness-adjustment procedures in such scenarios.

\subsection{Unexplored Issues in the Theory of Fair-RL}
In this section, we saw articles contributing mainly to the theory of fair-RL. Albeit, they advance the field by providing a mathematically-grounded understanding of the used methods, the following avenues are still to be explored.
\begin{enumerate}
    \item Fairness-performance trade-off: The fairness-performance trade-off is a fundamental question in fair-ML, and it has been discussed in several existing works \citep{pmlr-v81-menon18a, 10.5555/3327144.3327272}. The necessity of this trade-off has been contended in other forms of learning \citep{10.1145/3447548.3467326, 10.5555/3454287.3455691}. Indeed, a recent empirical study challenges the assumption that adhering to fairness constraints necessarily leads to a noticeable drop in accuracy \citep{Rodolfa2021}. However, this important matter is mostly not investigated in the theory of fair-RL yet  (except a few articles, e.g., \citep{10.5555/3327144.3327185,wang2019evolving, JMLR:v22:20-704, 10.1145/3287560.3287601}).
    \item Strictness of performance bounds: Almost all articles (barring a few, e.g., \citet{NIPS2016_eb163727}, \citet{pmlr-v139-wang21b}) do not prove a corresponding lower bound on the used performance measures. This makes it hard to determine the strictness of the performance bounds, i.e., how far they are from the best possible bounds. 
    \item Credible surveys from social sciences to determine the applicability of the fairness measures: 
    Most of the research does not make explicit use of the vast amount of surveys available in social sciences assessing the applicability of fairness notions for particular domains (see \citet{GajPechFAT2018} and references therein). These surveys could help RL researchers and practitioners to determine the most suitable fairness notions for the target domain. Furthermore, the testimony of these surveys will also justify the usage of the RL approaches employing the said fairness notions. This point is further discussed in the challenges given in Section \ref{Sec:CorerctBias} and \ref{sec:GroupOrInd}.  
\end{enumerate}

\section{Present challenges and future directions}
\label{sec:Challenges}
In this section, we describe major challenges in fair-RL research and discuss possible solutions to overcome them.

\subsection{Correcting Societal Biases}
\label{Sec:CorerctBias}
In this article, we reviewed a number of computational approaches aiming to alleviate societal biases. However, it is important to recognize and take into consideration the fact that the presence of biases in automated decisions is due to issues such as unequal access to resources and social conditioning \citep{SexBias1975,RePEc:oxp:obooks:9780198081692}. This is true for all forms of learning; nevertheless, it assumes additional relevance for RL methods as they traditionally learn from their own experiences without a knowledgable external supervisor. The absence of such a supervisor precludes the possibility to externally correct biases during the decision-making process.  

Since individuals should not be held responsible for the attributes they can not change or had no say in (i.e., sensitive attributes), the automated decisions, which affect the social benefits they receive and their prospects in life, should not depend upon those attributes. If RL algorithms are used to make decisions about the benefits whose allocation exhibits discrimination for some people owing to the attributes they had no say in, then it is reasonable that the algorithms should offset the existing discrimination due to such attributes. Of course, the obvious difficulty lies in determining which attributes that an individual has no say in. Education, at first glance, might seem like an attribute that an individual can choose. However, several studies show that the attributes that an individual has no say in (e.g., birth-place, race, caste) can impact the level of their education. For example, \citet{Jacobs2003} provide a comprehensive survey on education in the USA, which demonstrates that women are particularly disadvantaged with respect to the outcomes of education. This can be seen in academia in the USA as women make up only $26\%$ of full professors, $23\%$ of university presidents, and $14\%$ of presidents of doctoral degree-granting institutions \citep{WomenLead2009}. Similar gender disparity at higher echelons of positions is found in politics and business.
Moreover, this is despite the fact that women fare relatively well in access to education in the USA. Indeed, $57\%$ of all the college students in the USA are women \citep{WomenLead2009}. However, easy access to education for women is not a universal phenomenon, and discrimination is present in many parts of the world as confirmed by \citet[Chapter 3]{WorldWomen2015}. This leads to us a key insight that the level of discrimination varies with the domain (or even for different issues within a single domain as seen above) and the place of interest. Thus, while proposing fair-RL (and in general fair-ML) algorithms, credible surveys appraising the discrimination caused by the protected attributes in the target domain should be taken into consideration.   

Which fairness approaches should be used for tasks corresponding to a particular social benefit should also depend upon whether the benefit in question can be considered to be a basic human right. For domains like affordable housing, essential health-care, and basic education, fairness approaches that actively try to remove disparity caused due to protected attributes and provide benefits to all the individuals should be considered. However, for other domains which require qualifications not evenly distributed in the population, a justification could be made for relaxing these stipulations. At the same time, independent efforts could be made to diffuse the ability to have such qualifications evenly in the population.  

\subsection{Group Fairness or Individual Fairness?}
\label{sec:GroupOrInd}

While individual fairness and group fairness remain the most popular approaches in fair-RL, their applicability has been called into question. 
%
It has been argued that group fairness measures may miss unfairness against people as a result of their membership to multiple protected groups \cite{10.2307/1229039}, or groups which are not (yet) defined in anti-discrimination laws but may need protection \citep{Wachter2019}.
It has also been noted that many of the most promising group fairness constraints are mutually incompatible \cite{doi:10.1089/big.2016.0047}. If group fairness measures are applied to protected groups, such as race and gender, separately, they might permit unfairness for structured combinations of those groups (e.g., black women), also known as \textit{fairness gerrymandering} \cite{pmlr-v80-kearns18a}. Moreover, if group fairness is applied across many different combinations of protected characteristics, it does not scale well with a large number of groups and may lead to overfitting \cite{pmlr-v80-hebert-johnson18a}.

On the other hand, \citet{10.1145/3461702.3462621} present the following critiques of individual fairness --
\begin{enumerate*}[label=\roman*)]
\item counterexamples show that similar treatment guaranteed by individual fairness is insufficient to guarantee fairness, 
\item the used similarity metrics/arbiters may suffer from implicit systematic biases,
\item it cannot offer a substantive, non-circular definition of fairness because determining which features are task-relevant and thus apt for measuring similarity requires making moral judgments about what fairness constitutes, and
\item if incommensurable moral values are relevant for determining similarity for a task, similarity cannot be represented as a distance metric.
\end{enumerate*}

Furthermore, a recent work argues that the conflict between group fairness and individual fairness is more of an artifact of the blunt application of fairness measures, rather than a matter of conflicting principles \citep{10.1145/3351095.3372864}. \citet{pmlr-v28-zemel13} also suggest pairing individual fairness with statistical parity conditions. While the stated aim in \citet{8731591} is individual fairness, they claim that their approach indirectly improves group fairness because it reduces information on protected attributes. 

How to disentangle the apparent conflict between group fairness and individual fairness and improve their applicability remains an important challenge for fair-RL and fair-ML researchers and practitioners.

\subsection{Explainability in RL}
\label{sec:ExpRL}
Another important issue to consider is the opacity of many RL systems since, unlike other forms of learning, RL algorithms work using a trial-and-error method without human interaction. It has been shown that transparency not only increases users' trust \citep{10.1145/1378773.1378804} but also user's acceptance of a system \citep{10.1145/358916.358995}. The importance of explainability to RL systems is crucial from a legal perspective (see GDPR \cite{Goodman_2017}), and it also enhances their usefulness \citep{DBLP:journals/corr/abs-1901-00188}.  

The field of eXplainable AI (XAI) is aimed at verifying the fairness of ML systems by providing insights into how the machine learning model generates predictions.
XAI has mainly focused on supervised learning, but recent efforts in eXplainable Reinforcement Learning (XRL) are specific to reinforcement learning. \citet{verma2018programmatically} introduce an inherently interpretable model framework, called Programmatically Interpretable Reinforcement Learning (PIRL), that represents policies using a high-level, human-readable programming language. This is achieved by limiting the set of target policies to conform to a policy sketch: a grammar of expressions over atoms and sequences of atoms, essentially similar to regularization.
\citet{liu2018toward} introduce an XRL post-hoc explainability approach that mimics the model's Q-function using Linear Model U-Trees (LMUTs): simple and understandable models designed to approximate continuous functions. 
\citet{DBLP:journals/corr/abs-1712-07294} propose a framework for multi-task reinforcement learning which is capable of composing hierarchical plans in an interpretable manner.  \citet{DBLP:journals/corr/abs-1905-10958} present a method to generate explanations of the behavior of RL agents based on counterfactual analysis of the causal model.
For more details on explainable reinforcement learning, we refer to the survey by \citet{puiutta2020explainable}. 
  
Notwithstanding these works, explainability for the intended target audience is often neglected in the development of RL methods \citep{10.1145/3173574.3174156} and remains a credible challenge for the fair-RL community. It is also important to cater the form and presentation of explanations to the needs of the target audience as it has been shown that human users favor some forms of explanations over others \citep{MILLER20191}. 

\section{Concluding remarks}
\label{sec:conclusion}
Fairness in reinforcement learning is a rapidly-growing emerging field, where newer solutions are revealing newer challenges, and there are immense opportunities to take this discipline forward theoretically as well as empirically. At this juncture, we present an extensive survey covering the full breadth of fair-RL. The organization of this survey into various aspects of RL and applications of fair-RL is with the intention of aiding researchers and practitioners to gain a thorough understanding of the field. Towards the end, we also discuss a few of the most important challenges, which we hope will lead to productive conversations and advance the field of fair-RL.

\bibliographystyle{ACM-Reference-Format}
\bibliography{references}

\appendix
\section{List of Symbols}
\label{Appd:SymbolTable}
\printnomenclature
\section{Theory of Fair-RL}
\label{Appd:TheoryFairRL}

\citet{DBLP:journals/corr/JosephKMNR16} consider contextual bandits in which each arm (i.e., individual) is represented by a vector of $\dimension$ 
\nomenclature{$\dimension$}{Dimension of a feature vector}%
features.
For this setting, they propose an algorithm called \Alg{RidgeFair}\textsubscript{m} with a regret bound of $\tilde{O}(\dimension |\set{A}|^2\sqrt{T})$. Secondly, they provide an algorithm for infinite bandits.  Recall that, in this setting the learner sees a convex set of arms contained in a ball of radius $\radius$.
\nomenclature{$\radius$}{Radius of a ball}%
The provided algorithm has a regret bound of $\tilde{O}\left(\frac{\radius^6 \subgaussian^2}{\kappa^2\lambda^2 \Delta_{gap}^2} \right)$ where $\lambda$ measures how quickly an agent can learn from uniformly random actions, $\kappa = 1 - \radius \sqrt{\frac{2 \log{(2dT/\delta)}}{T\lambda}}$, and $\Delta_{gap}$ is a lower bound on the difference in expected reward of the optimal action and any other extremal action in a convex set seen by the learner.
\nomenclature{$\lambda$}{Measure of how quickly an agent can learn from uniformly random actions \citep{DBLP:journals/corr/JosephKMNR16}}%
\nomenclature{$\delta$}{Confidence in a Probably Approximate Correct(PAC) bound}
\nomenclature{$\epsilon$}{Accuracy in a Probably Approximate Correct(PAC) bound}
\nomenclature{$\kappa$}{$1 - \radius \sqrt{\frac{2 \log{(2dT/\delta)}}{T\lambda}}$ \citep{DBLP:journals/corr/JosephKMNR16}}%
\nomenclature{$\Delta_{gap}$}{Lower bound on the difference in expected reward of the optimal action and any other extremal action in a convex set seen by the learner\citep{DBLP:journals/corr/JosephKMNR16}}

\citet{1112306} propose using individual fairness in stochastic bandits. To that end, they impose: 
\begin{enumerate*}[label=\roman*)]
    \item \textit{smooth fairness} constraint - two arms with similar distribution should be selected with similar probability, and 
    \item \textit{calibrated fairness} constraint - sample each arm with probability equal to its reward being the greatest.
\end{enumerate*}
The authors elucidate how these two constraints address the first two critiques of meritocratic fairness presented earlier. As it turns out, this notion of exact calibrated fairness is not possible to achieve during the learning phase, hence the authors aim to minimize expected cumulative amount by which the algorithm violates calibrated fairness constraint across time-steps. They propose a variation of Thompson sampling adhering to smooth fairness for total variation distance and prove a $\tilde{O}\left((|\set{A}|T)^{2/3}\right)$ upper bound on fairness regret. Furthermore, they also modify their algorithm for dueling bandits and prove a $\tilde{O}\left(|\set{A}|^{4/3}T^{2/3}\right)$ bound on fairness regret.

\citet{10.5555/3327144.3327185} consider individual fairness constraints governed by a Mahalanobis similarity metric in linear contextual bandits. The authors assume that the algorithm has access to an oracle that can ascertain what it means to be fair, but cannot explicitly enunciate the fairness metric. The fairness constraint necessitates that the difference between the probabilities with which any two actions are taken is bounded by the distance between their contexts. After the learner makes its decisions, it receives feedback from the oracle specifying for which pairs of contexts the fairness constraint was violated. The goal is to guarantee near-optimal regret (with respect to the best fair policy), while violating the fairness constraints as infrequently as possible. They propose an algorithm, which --
\begin{enumerate*}[label=\roman*)]
\item achieves regret of $\tilde{O}\left( |\set{A}|^2 \dimension^2 \log{(T)} + \dimension \sqrt{T} \right)$ with a high probability, where $d$ is the dimension of the context vectors, and
\item violates the fairness constraint by more than $\epsilon$ on at most $O\left( |\actions|^2 \dimension^2 \log{(\dimension/\epsilon)}\right)$ steps.
\end{enumerate*}
The above guarantees are obtained by setting $\epsilon = O(1/T)$. Other trade-offs between regret and fairness violations are also possible.

In the setting explored by \citet{DBLP:journals/corr/abs-1912-03802}, the reward for each arm belonging to protected groups is additionally corrupted with a bias term. The performance measure is cumulative regret ignoring the bias terms. The authors propose an algorithm called \Alg{GroupFairTopInterval} which can accommodate both their definitions of group fairness -- demographic parity and proportional parity. 

\citet{pmlr-v130-wen21a} develop fair decision-making policies in discounted MDPs. They consider the episodic case where the system is reset after a fixed number of steps. They distinguish between the quality of outcomes for the decision-maker and the quality of outcomes for an individual with the latter being termed as \textit{individual rewards}. Individual rewards are assumed to be state independent. The goal is to learn an optimal policy that does not favor the majority sub-population over the minority sub-population in terms of individual rewards. In particular, their goal works with demographic parity and equal opportunity \citep{NIPS2016_6374}. The performance bound is the expected regret in terms of cumulative reward for the decision-maker over total number of episodes $\numEp$.
\nomenclature{$\numEp$}{Number of episodes}%
They use explore-then-commit strategy \citep{lattimore_szepesvari_2020} where the number of exploration episodes is computed using the number of steps in an episode, number of states, an upper bound on the reward, and other problem-dependant factors. They guarantee that the above strategy satisfies demographic parity with arbitrary tolerance $\epsilon$ and prove $O\left( \left( N^{2/3} + 1/\epsilon^2 \right) \log{(1/\delta)}\right)$ upper bound on the expected regret with probability $1-\delta$.
\nomenclature{$\epsilon$}{Tolerance w.r.t. demographic parity in \citet{pmlr-v130-wen21a}}%
A noteworthy caveat for this bound is that the reference optimal policy is required to satisfy a stricter level of fairness.

\citet{8737461} explore a variation of classical stochastic bandits called combinatorial sleeping bandits in which multiple arms (upto $\maxarms$)
can be simultaneously played and an arm could sometimes be ``sleeping" (i.e., unavailable). The fairness criterion is to have a minimum selection fraction for each individual arm. They propose a variation of UCB algorithm \cite{UCB} called \Alg{Learning with Fairness Guarantee (LFG)} which 
satisfies asymptotic fairness guarantee. As a performance measure, they use time-average regret which is the difference between the expected per-step reward of the optimal policy satisfying the fairness criterion and that of the considered algorithm. They provide an \textit{instance-independent}\footnote{See definition 7 in \citet{MAL-068} } upper bound on the time-average regret of 
$\frac{|\set{A}|}{2\eta} + O\left( \frac{\sqrt{\maxarms |\actions|T \log{T}} + |\actions|}{T} \right)$ for \Alg{LFG},
where $\eta$ is a design parameter which controls the priority of meeting the fairness requirement. 
\nomenclature{$\eta$}{Design parameter which controls the priority of meeting the fairness requirement in \citet{8737461}}%
The authors enlist real-time scheduling in wireless networks, ad placement in online advertising systems, and task assignment in crowd-sourcing platforms as practical applications and present experimental results on simulations showing competent performance of \Alg{LFG}.

\citet{JMLR:v22:20-704} consider stochastic bandits where the minimum selection fraction of each arm is limited to $\left[0,\frac{1}{|\set{A}| -1}\right)$. The performance measure is called fairness-aware regret in which the expected cumulative reward of the considered algorithm is compared to that of an optimal policy satisfying the fairness constraint.
They propose a meta-algorithm called \Alg{FAIR-Learn} which either selects the arm straying the most from the fairness constraint if such an arm exists or selects an arm according to the MAB algorithm used as a black-box. The authors prove that the fairness guarantees hold uniformly over time and provide a $O(\sqrt{T\log{T}})$ bound on the fairness-awareness regret.

\citet{pmlr-v139-wang21b} aim to achieve merit-based fairness of exposure to the items while optimizing utility to the users in classical stochastic bandits and stochastic linear bandits. The fairness constraint requires that each arm receives an amount of exposure proportional to its merit, where merit is quantified through an application-dependent merit function given to the algorithm. They characterize the merit function --
\begin{enumerate*}[label=\roman*)]
    \item to have some minimum positive constant merit, and
    \item to be Lipschitz continuous. 
\end{enumerate*}  
The necessity of these two conditions is established by proving a $O(T)$ lower bound on expected fairness regret when either of the conditions is relaxed. They distinguish between \textit{reward regret} -- \ the gap between the expected cumulative reward of a policy and that of the optimal fair policy, and \textit{fairness regret} -- \ the cumulative L1 distance between a policy and the optimal fair policy. For stochastic bandits, they propose a frequentist algorithm called \Alg{FairX-UCB} and a Bayesian algorithm called \Alg{FairX-TS}. For \Alg{FairX-UCB}, they prove a high-probability bound of $\tilde{O}\left(LipschitzConstant \cdot \sqrt{|\actions|T}/(MinimumMerit) \right)$ on fairness regret and a high-probability bound of $\tilde{O}\left(\sqrt{|\actions|T}\right)$ on reward regret. For \Alg{FairX-TS}, they prove the same respective bounds for Bayesian fairness regret and Bayesian reward regret. They further extend the algorithms to \Alg{FairX-LinUCB} and \Alg{FairX-LinTS} for stochastic linear bandits. For both algorithms, they prove a bound of $\tilde{O}\left( \dimension \sqrt{T}\right)$ on the reward regret (resp. Bayesian reward regret) where $\dimension$ is the dimensionality of the context vectors. In addition, they prove a high-probability bound of $\tilde{O}\left( LipschitzConstant \dimension \sqrt{T}/(MinimumMerit)\right)$ on fairness regret for \Alg{FairX-LinUCB}, and for \Alg{FairX-LinTS}, they prove a bound of $\tilde{O}\left( LipschitzConstant \sqrt{\dimension T}/(MinimumMerit)\right)$ on Bayesian fairness regret. Furthermore, the experimental results on synthetic data and on real-world data from ``Yahoo! Today" Module \citep{10.1145/1772690.1772758} show that the proposed algorithms provide better fairness in terms of exposure to the items.

\citet{DBLP:journals/corr/abs-2111-14387} provide a RL solution for fair and efficient allocation of a divisible resource among a population with distinct preferences. They model this problem using adversarial bandits where the learner chooses the next fraction of the resource to be distributed and then an adversary specifies each individual's (i.e. arm's) valuation of that fraction.  The agent allocates the fraction to one of the arms and receives a reward dependent on the selected arm's valuation.  Fairness measure of a policy is the difference between the highest
accumulated valuation among all the arms and the lowest. Fairness-regret of a policy is the difference between the fairness measure of an optimum offline policy and the said policy. They propose an algorithm based on \Alg{EXP3} \citep{10.1137/S0097539701398375} which attains sub-linear bounds for both cumulative regret and fairness-regret if the adversary is restrained to only assign certain kind of evaluations.  

\end{document}